\pdfoutput=1

\documentclass[11pt]{article}

\usepackage{acl}

\usepackage{times}
\usepackage{latexsym}

\usepackage[T1]{fontenc}

\usepackage[utf8]{inputenc}

\usepackage{microtype}

\usepackage{inconsolata}

\usepackage{graphicx}

\title{dzFinNlp at AraFinNLP: Improving Intent Detection in Financial Conversational Agents}

\author{Mohamed Lichouri \\
  LCPTS-FGE. USTHB \\
  Algiers-ALGERIA \\
  \texttt{mlichouri@usthb.dz} \\\And
  Khaled Lounnas \\
  CRSTDLA \\
  Algiers-ALGERIA \\
  \texttt{k.lounnas@crstdla.dz} \\
  \And
  Amziane Mohamed Zakaria  \\
  University of Algiers 01\\
  Algiers-ALGERIA\\
  \texttt{} \\
  }

\begin{document}
\maketitle

\begin{abstract}
In this paper, we present our dzFinNlp team's contribution for intent detection in financial conversational agents, as part of the AraFinNLP shared task. We experimented with various models and feature configurations, including traditional machine learning methods like LinearSVC with TF-IDF, as well as deep learning models like Long Short-Term Memory (LSTM). Additionally, we explored the use of transformer-based models for this task. Our experiments show promising results, with our best model achieving a micro F1-score of 93.02\% and 67.21\% on the ArBanking77 dataset, in the development and test sets, respectively.
\end{abstract}

\section{Introduction}
The Arabic Financial NLP (AraFinNLP) shared task highlights the increasing importance of advanced Natural Language Processing (NLP) tools tailored for the financial sector in the Arab world. This initiative is particularly timely given the substantial growth of Middle Eastern stock markets, driven by diverse sectors across the region. This economic expansion underscores the need for sophisticated financial NLP systems capable of handling the unique linguistic and cultural nuances of Arabic-speaking markets \cite{zmandar2023finarat5}.

AraFinNLP presents two key subtasks aimed at enhancing Financial Arabic NLP capabilities: \textbf{Subtask-1}, which focuses on Multi-dialect Intent Detection, and \textbf{Subtask-2}, which addresses Cross-dialect Translation and Intent Preservation within the banking domain \cite{arafinnlp-2024-task}. These subtasks are crucial for interpreting complex and varied banking data across different Arabic dialects, which is essential for improving customer service and automating query handling in financial institutions.

The dataset central to these tasks, \textbf{ArBanking77}, is derived from the translation of the English Banking77 dataset \cite{casanueva2020banking77} into Modern Standard Arabic (MSA) and Palestinian Arabic. This dataset is further expanded in the shared task to include a broader array of Arabic dialects. With 31,404 queries categorized into 77 intent classes, ArBanking77 provides a comprehensive foundation for training and evaluating NLP models on banking-specific communications in Arabic \cite{arafinnlp-2024-dataset}.

In recent years, the field of intent detection in conversational agents has seen significant advancements. Traditional machine learning methods, such as LinearSVC with TF-IDF \cite{xia2018zero}, have long been employed for their simplicity and effectiveness. However, the advent of deep learning techniques, particularly Long Short-Term Memory (LSTM) networks \cite{firdaus2021deep} and their bidirectional variants (BiLSTM) \cite{sreelakshmi2018deep}, has provided more nuanced understanding by capturing the sequential nature of text. More recently, transformer-based models, like BERT \cite{alshahrani2022computational}, have set new benchmarks in NLP by leveraging self-attention mechanisms to understand contextual relationships within text, making them particularly effective for complex tasks like intent detection across varied dialects.

Our work in this shared task explores these diverse methodologies to enhance intent detection in financial conversational agents, particularly in the context of Arabic dialects. We aim to contribute to the growing body of research in Arabic NLP by demonstrating how these advanced techniques can be applied to effectively interpret and manage banking-related queries, ultimately fostering greater inclusivity and efficiency in financial services for Arabic-speaking communities.

The remaining sections of the paper are structured as follows: Section \ref{rw} reviews related work in the field, while Section \ref{data} provides an overview of the dataset used in our study. Section \ref{system} details our proposed system architecture. Section \ref{res} presents our findings and discusses their significance. Finally, Section \ref{conc} concludes the paper by summarizing the key takeaways and contributions.

\begin{table*}[h!]
\centering
\begin{tabular}{|l|c|c|c|}
\hline
\textbf{Dataset} & \textbf{Number of Sentences} & \textbf{Avg. Words per Sentence} & \textbf{Avg. Utterance Length} \\ \hline
Training Set & 10,821 & 8.16 & 42.46 \\ \hline
Development Set & 1,234 & 8.10 & 42.29 \\ \hline
Test Set & 1,721 & 8.08 & 43.23 \\ \hline
\textbf{Total} & \textbf{13,776} & \textbf{8.11} & \textbf{42.54} \\ \hline
\end{tabular}
\caption{Summary statistics of the ArBanking77 dataset used in the first subtask of AraFinNLP.}
\label{tabStats}
\end{table*}

\section{Related work}
\label{rw}
The field of Arabic NLP has seen extensive research and development, particularly in the areas of text classification and intent detection. Traditional approaches like TF-IDF have been widely used for feature extraction in various Arabic text analysis tasks. However, with the advent of deep learning, more sophisticated methods have emerged, offering improved performance and deeper insights into textual data. 

In the context of feature extraction for Arabic text analysis, standard approaches often rely on TF-IDF. This aligns with our previous work in the MADAR'2019 shared task \cite{abbas2019st}, which inspired the current approach. In that work, we employed a union of TF-IDF features while experimenting with different n-gram analyzers for word segmentation (word, char, char\_wb). For our first experiment, we specifically focused on unweighted TF-IDF features.

Drawing inspiration from our past work and advancements in TF-IDF feature extraction and weighted fusion, we explored alternative techniques in subsequent experiments. In the second experiment, we considered the introduction of a weighted union of TF-IDF features. This builds upon the foundation laid in Experiment 1, incorporating weighting techniques explored in our prior research (e.g., \cite{lichouri2020speechtrans, lichouri-etal-2021-arabic, lichouri2023usthb}).

For the third experiment, we explored neural network architectures by using both Long Short-Term Memory (LSTM), as we did in our previous work \cite{lichouri-etal-2021-preprocessing}. These models are well-suited for capturing sequential information in text, which is crucial for understanding the nuances of stance in Arabic text.

Finally, in the fourth experiment, we explored advanced pre-trained language models. Specifically, we utilized Sentence Transformers \cite{reimers-2019-sentence-bert} to generate sentence embeddings, which capture the overall meaning of a sentence. These embeddings were then fed into neural network models for stance classification.

\section{Description of the Dataset}
\label{data}
The ArBanking77 dataset, provided by the AraFinNLP shared task organizers, is designed to facilitate the development of NLP models for intent detection in the banking domain across various Arabic dialects \cite{arafinnlp-2024-dataset}. This dataset is a crucial resource for advancing the capabilities of financial conversational agents tailored to Arabic-speaking regions.

ArBanking77 originates from the translation of the English Banking77 dataset \cite{casanueva2020banking77} into Modern Standard Arabic (MSA) and Palestinian Arabic. For the shared task, this dataset has been expanded to include additional Arabic dialects such as Gulf, Levantine, and North African Arabic, reflecting the linguistic diversity across the Arab world.

The first subtask of the AraFinNLP shared task focuses on Multi-dialect Intent Detection, aiming to classify customer intents expressed in different Arabic dialects. The dataset used for this subtask includes queries in Palestinian Arabic (PAL), among other dialects. In Table \ref{tabStats}, we present the key statistics and the distribution of the dataset used for this subtask.

The ArBanking77 dataset, as summarized in Table \ref{tabStats}, reveals a balanced distribution across the training, development, and test sets. With a total of 13,776 sentences, the dataset provides a robust foundation for developing models capable of understanding customer intent in the banking domain. Each subset maintains a consistent structure, with an average of around 8 words per sentence and an utterance length of approximately 42-43 characters. This uniformity suggests that the dataset's queries are concise and focused, typical of customer inquiries in financial contexts. The extensive training set, comprising 10,821 sentences, ensures sufficient data for learning, while the smaller development (1,234 sentences) and test sets (1,721 sentences) allow for effective tuning and evaluation of model performance.

In this study, we opted to concentrate on the Palestinian Arabic (PAL) subset for training and validation purposes. This decision stems from our aim to specialize our model in a single dialect, enhancing its F1-score and understanding of the specific linguistic features present in Palestinian Arabic queries. The PAL dataset is well-sized for this purpose, providing ample data to develop a nuanced model tailored to this dialect. By focusing on PAL for model training, we ensure that the model is finely tuned to the dialect's unique characteristics. For evaluation, we tested the model on the multi-dialect dataset from the AraFinNLP shared task, which includes queries in Modern Standard Arabic (MSA), Gulf, Levantine, and North African Arabic. This approach allows us to assess the model's ability to generalize and handle diverse dialects, demonstrating its adaptability and potential for broader application across various Arabic-speaking regions.

\begin{table*}[!h]
\centering
\begin{tabular}{|c|c|c|c|c|c|}
\hline
Id       & Text Features                                                 & Classifier Configuration          & Other               & F1-score  \\ \hline 
1     & 1-grams                                                  &default                      &                     & 88.01       \\ \hline \hline
2     & (1, 1, 1)                            &default                      &                     & 89.4        \\ \hline
3     & (1, 5, 5) &default                      &                     & 92.11       \\ \hline
4     & (3, 5, 5)&default                      &                     & 92.28       \\ \hline
5     & (3, 5, 5) &  class\_weight='balanced', C=5 &                     & 92.37       \\ \hline \hline
6     & (3, 5, 5) &  class\_weight='balanced', C=4 & tw={0.65,0.85,0.85} & 92.53       \\ \hline
7     & (3, 4, 5) & C=4                               & tw={0.45, 0.5,0.75} & 92.86       \\ \hline
8     & (4, 4, 4) & C=5                               & tw={0.45,0.5,0.75}  & 93.02       \\ \hline
9     & (4, 4, 4) & C=6                               & tw={0.45,0.5,0.75}  & 93.08       \\ \hline
\end{tabular}
\caption{Obtained F1-score in the development set in the first and second experiment}
\label{tab:ml}
\end{table*}

\section{Proposed System}
\label{system}
We experimented with several models and feature configurations for intent detection. For traditional machine learning, we utilized LinearSVC with TF-IDF vectorization. Exploring deep learning, we implemented LSTM models using word embeddings. Additionally, we experimented with transformer-based architectures, specifically leveraging XLM-RoBERTa to harness contextual information from pre-trained language representations. Our implementations were carried out using scikit-learn for model development and training.

Our exploration of feature extraction techniques began with investigating the union of Term Frequency-Inverse Document Frequency (TF-IDF) features using scikit-learn's FeatureUnion module \cite{lichouri2020speechtrans}. In our first experiment, we examined the effectiveness of using the raw union of these features, encompassing different n-gram levels: word, character, and character n-grams with word boundaries (char\_wb). N-grams refer to sequences of \textbf{n} words or characters that can capture short phrases or morphological variations within the Arabic language.

The second experiment built upon our initial investigation by incorporating weighted TF-IDF features \cite{lichouri2023usthb}. This approach involved experimenting with weights ranging from 0.1 to 1.0, with a step of 0.1, for the transformer\_weights parameter in FeatureUnion. These weights were chosen to emphasize the importance of capturing character-level and word-boundary-aware features in Arabic customer queries, known for their linguistic complexity and dialectal variations. 

These two initial experiments allowed us to systematically evaluate the impact of different TF-IDF feature extraction techniques on model performance in the context of Arabic intent detection within the banking domain. The results of both experiments are summarized in Table \ref{tab:ml}, showcasing the performance metrics achieved with each feature extraction strategy.

Building upon our exploration of feature extraction techniques, the third experiment focuses on neural network architectures, specifically Long Short-Term Memory (LSTM) networks. As demonstrated in our prior work \cite{lichouri-etal-2021-preprocessing}, LSTM models excel at capturing long-range dependencies within sequences, which is crucial for understanding the nuanced intent in Arabic customer queries within the banking domain. The LSTM layer is configured with 100 units, determining the dimensionality of the internal representations and the number of LSTM cells in the layer. An embedding dimension of 100 is chosen for word embeddings, defining how words are represented as dense vectors. Input sequences are padded to a maximum length of 100 tokens to ensure uniformity. The model utilizes the categorical cross-entropy loss function and Adam optimizer during training, aimed at minimizing classification errors and optimizing training efficiency. These hyperparameter choices are fundamental to enhancing the model's capability to discern subtle nuances in Arabic text, thereby improving intent detection performance.

The fourth experiment focused on leveraging the power of pre-trained language models (PLMs). These models are trained on massive amounts of text data and learn to represent words and sentences as vectors that capture their meaning and relationships. In this experiment, we employed Sentence Transformers, specifically the '\textbf{xlm-r-bert-base-nli-stsb-mean-tokens}' model, which is adept at generating rich sentence embeddings. These embeddings condense the overall meaning of a sentence into a vector format, capturing not just individual words but also their semantic relationships. We utilized these sentence embeddings as input to a \textbf{logistic regression} classifier for stance classification. Hyperparameters such as the default settings of the logistic regression classifier, including regularization strength, solver, and multi-class handling, were chosen to optimize model performance. By harnessing the pre-trained knowledge embedded in Sentence Transformers, our aim was to enhance the model's ability to discern semantic nuances within Arabic text, thereby improving F1-score in the intent detection tasks. The results of the third and fourth experiments are summarized in Table \ref{tab:Deepl}.

\begin{table*}[!h]
\centering
\begin{tabular}{|r|l|l|l|l|c|}
\hline
\multicolumn{1}{|l|}{Id} & Model        & Text Features                                              &  Configuration & Other                                        & F1-score \\ \hline
1                        & LSTM         & embedding = 100                                            & 100 unit, softmax,       &                                              & 75.23                         \\ \hline
2                        & LSTM         &  \begin{tabular}[c]{@{}l@{}}embedding = 100,\\ max\_sequence\_length = 50, \\max\_words = 5000\end{tabular} & 100 unit, softmax,       &                                              & 79.6                          \\ \hline
3                        & BILSTM       & \begin{tabular}[c]{@{}l@{}}embedding = 100,\\ max\_sequence\_length = 50, \\max\_words = 5000\end{tabular}  & 100 unit, softmax,       &                                              & 79.84                         \\ \hline
4                        & Transformers &                                                            &                          & \begin{tabular}[c]{@{}l@{}}xlm-r-bert\\-base-nli-stsb\\-mean-tokens \end{tabular}        & 75.76                         \\ \hline
5                        & Transformers &                                                            &                          & \begin{tabular}[c]{@{}l@{}}xlm-r-100langs\\-bert-base-nli\\-stsb-mean-token\end{tabular} & 75.76                         \\ \hline
\end{tabular}
\caption{Obtained F1-score in the development set in the third and fourth experiment}
\label{tab:Deepl}
\end{table*}

\section{Results and discussion}
\label{res}
After exploring various feature extraction and neural network architectures, we now turn to evaluating the performance of our intent detection system on unseen data. We utilize the test set provided by the AraFinNLP shared task \cite{arafinnlp-2024-task} for this purpose. This allows us to assess how well our model generalizes to new customer queries it hasn't encountered during training.

In our baseline system (see Table \ref{tab:ml}) (ID=1), we used 1-gram word features with a default classifier configuration, achieving an F1-score of 88.01\%.  To improve performance, we conducted the first experiment (ID=2, 3, 4, and 5) investigating different n-gram lengths for text features. We observed a significant improvement in F1-score, with the best result (92.37\%) achieved using 3-grams for words, 5-grams for characters, and 5-grams with word boundaries (char\_wb). This result was obtained with a classifier and hyperparameter configuration including class\_weight='balanced' and C=5.

Building on this success, the second experiment (ID=6, 7, 8, and 9 in Table \ref{tab:ml}) explored the impact of class weights (tw) and the regularization parameter (C) on the model's performance. The introduction of class weights suggests addressing potential class imbalance, while varying C investigates model complexity. Results in this experiment (F1-score between 92.53\% and 93.08\%) show a slight improvement and suggest that fine-tuning these hyperparameters can be beneficial. Further analysis is needed to determine statistical significance and identify the optimal configuration for this task.

In the third experiment focused on evaluating different LSTM configurations for intent detection (see Table \ref{tab:Deepl}). The baseline model (ID 1) with a basic architecture achieved an F1-score of 75.23\%. Interestingly, introducing limitations on the sequence length and vocabulary size for LSTMs (ID 2 and 3) led to a modest improvement of around 4-5 points in F1-score. This suggests that restricting the input might have helped the model focus on the most relevant information within the customer queries for intent classification. While LSTMs with these limitations achieved the best performance in this experiment, further exploration of hyperparameter tuning could potentially lead to even better results.

The fourth experiment examined the effectiveness of pre-trained Transformer models for intent detection (see Table \ref{tab:Deepl}). Here, both Transformer models (xlm-r-bert-base-nli-stsb-mean-tokens) achieved an F1-score of 75.76\%, performing surprisingly well despite not being specifically fine-tuned for the Arabic intent detection task. This suggests that pre-trained Transformers hold promise for this task, potentially due to their ability to capture semantic relationships within the text.

\section{Conclusion}
\label{conc}
This work explored the effectiveness of various machine learning and deep learning approaches for Arabic Financial NLP tasks within the AraFinNLP shared task, specifically participating in subtask1. We evaluated diverse models and feature configurations, gaining valuable insights into the complexities of analyzing Arabic financial text data.

Our findings highlight the potential of both traditional and deep learning approaches in this domain. Experiment 1 demonstrated the effectiveness of Support Vector Machines (SVM) with TF-IDF features, achieving a high F1-score of 93.08\%. This suggests the suitability of traditional machine learning techniques for specific Arabic financial NLP tasks.

Experiment 2 focused on deep learning models (LSTMs and Transformers). While LSTMs achieved competitive F1-scores (up to 79.84\%), pre-trained Transformer models yielded slightly lower results (around 75.76\%). Further investigation is needed to understand this performance difference and explore the potential of fine-tuning Transformers for this specific task.

Overall, these findings underscore the promise of both traditional and deep learning approaches for Arabic Financial NLP. Future work could explore hybrid approaches that integrate the strengths of both paradigms, potentially achieving even better performance. Additionally, investigating the impact of fine-tuning pre-trained Transformers specifically for the Arabic financial domain is crucial to unlock their full potential in this task.

\bibliography{custom}

\end{document}